\pdfoutput=1

\documentclass[11pt]{article}

\usepackage[review]{EMNLP2023}

\usepackage{times}
\usepackage{latexsym}
\usepackage{graphicx}
\usepackage{amsmath}
\usepackage{algorithm}
\usepackage{algorithmic}
\usepackage{booktabs}
\usepackage{float}

\usepackage{makecell}

\usepackage{appendix}

\usepackage[T1]{fontenc}

\usepackage[utf8]{inputenc}

\usepackage{microtype}

\usepackage{inconsolata}

\title{Crafting Efficient Fine-Tuning Strategies for Large Language Models}

\author{Michael Oliver \\
  Huski Inc \\
  \texttt{michael.o@huski.ai} \\\And
  Guan Wang \\
  Huski Inc \\
  \texttt{guan.w@huski.ai} \\}

\begin{document}
\maketitle
\begin{abstract}

This paper addresses the challenges of efficiently fine-tuning large language models (LLMs) by exploring data efficiency and hyperparameter optimization. We investigate the minimum data required for effective fine-tuning and propose a novel hyperparameter optimization method that leverages early-stage model performance. Our experiments demonstrate that fine-tuning with as few as 200 samples can improve model accuracy from 70\% to 88\% in a product attribute extraction task. We identify a saturation point of approximately 6,500 samples, beyond which additional data yields diminishing returns. Our proposed bayesian hyperparameter optimization method, which evaluates models at 20\% of total training time, correlates strongly with final model performance, with 4 out of 5 top early-stage models remaining in the top 5 at completion. This approach led to a 2\% improvement in accuracy over baseline models when evaluated on an independent test set. These findings offer actionable insights for practitioners, potentially reducing computational load and dependency on extensive datasets while enhancing overall performance of fine-tuned LLMs.

\end{abstract}

\section{Introduction}

Large Language Models (LLMs) have revolutionized the field of natural language processing. However, fine-tuning these models for specific applications remains a critical challenge, often requiring substantial computational resources and expert knowledge. This paper addresses these challenges by focusing on two key aspects of the fine-tuning process: data efficiency and hyperparameter optimization. Data efficiency is a critical consideration in fine-tuning, as the amount of labeled data required can be a limiting factor in many real-world scenarios. Collecting and annotating large datasets is time-consuming and expensive, and in some cases, such as in specialized domains or low-resource languages, extensive data may not be available. Therefore, understanding the minimum amount of data needed to achieve effective fine-tuning is crucial for optimizing resource allocation and reducing the burden of data acquisition.

Our experiments investigate the relationship between data volume and fine-tuning performance, seeking to identify a range where the model achieves strong results with the least amount of data.
By systematically varying the size of the fine-tuning dataset, we aim to provide insights into the diminishing returns of additional data beyond a certain threshold. These findings can guide practitioners in making informed decisions about data collection and allocation, ultimately leading to more efficient fine-tuning processes.

In addition to data efficiency, hyperparameter optimization plays a vital role in the success of fine-tuning. Hyperparameters, such as batch size, learning rate, and epochs, have a significant impact on the model's performance and convergence speed. However, finding the optimal combination of hyperparameters is a complex and time-consuming task, often requiring extensive trial and error or relying on heuristics and default values.

 
To streamline this process, we propose the use of Bayesian optimization techniques for hyperparameter search. Bayesian optimization offers a principled approach to exploring the hyperparameter space by balancing exploration and exploitation. By leveraging the information gained from previous evaluations, it intelligently suggests promising hyperparameter configurations to evaluate next. This approach allows for a more efficient search, reducing the number of computationally expensive training runs required to find optimal settings.

%

We study these two questions in the context of developing a solution to task of extracting specific information from web pages from a variety of e-commerce sites. For example, we seek to extract information such as product titles, descriptions, and prices from product pages, and contact information and social media links from contact pages.

The complexity of the task arises from the variability in page formats across different sellers and the need to adapt models to accurately capture this diversity with limited labeled data. This setting serves as a practical demonstration of our methods for exploring data efficiency and optimizing hyperparameters, showcasing their applicability in a real-world, domain-specific scenario.

\section{Related Work}



LLM fine-tuning has drawn tremendous research interest recently. Some research focuses on efficient fine-tuning methodologies, e.g., PEFT~\citet{ding2023peft} (Parameter-Efficient Fine-Tuning) and LORA (Low Rank Adaption)~\citet{hu2021lora}, while others focuses on RHLF data construction policies, e.g., \citet{tajwar2024preference}. However, to the best of our knowledge, there is little research being done for studying the bayesian optimization on hyperparameter selection for LLM fine-tuning. ~\citet{jin2023rethinkinglearningratetuning} proposed a learning rate adjustment strategy named LRBench++ that dynamically update learning rate among iterations.
Our work combines a systematic approach to searching hyperparameter combinations, including learning rate, batch size, LORA rank, etc. Our optimization target is to improve the model accuracy on the real testing data rather than training loss. 

Systematic and effective hyperparameter searching for LLM \emph{pretraining} is another related field. The primary motivation in this context is to identify optimal hyperparameter settings using small-scale training experiments, which are feasible on a limited number of GPUs with much less data compared to full-scale operations. These experiments aim to derive hyperparameter configurations that can be effectively applied in real pretraining tasks involving tens of thousands of GPUs and potentially lasting many months. This approach is designed to optimize resource allocation, ensuring that the extensive and costly resources involved in large-scale pretraining are used efficiently. The findings of \citet{yang2022tensor} underscore the practical utility of this approach, demonstrating how even minimal initial experiments can inform the scalable deployment of LLMs in ways that conserve computational effort while maximizing performance outcomes across extensive neural network architectures. This method aligns with the broader objective of making large-scale machine learning operations more sustainable and efficient, particularly in scenarios where computational resources are a significant bottleneck.
As what is going to be shown in Section 5, we also emphasized the intuition that if a set of hyperparameters yields better early testing accuracies it would be likely to produce better final accuracies. And using such intuition we could design the fine-tuning procedure more compute-efficient. 

Automatic hyperparameter search strategy has been a long-standing hot topic in machine learning community. \citet{wu2021frugaloptimization} proposed a local search approach that integrates a randomized direct-search method to effectively balance validation loss minimization with computational cost constraints. \citet{wang2021hpoblend} further combined this local search with a Bayesian optimization approach to form a "blend search" which is more robust for both tree-based models as well as neural networks. 
AutoLRS by \citet{jin2021autolrsautomaticlearningrateschedule} introduces a novel approach to dynamically adjust learning rates using Bayesian Optimization, illustrating significant improvements in training efficiency across several architectures, including ResNet-50 and BERT. This method highlights the potential of adaptive learning rate schedules to reduce computational overhead and improve model generalization without extensive manual hyperparameter tuning. However, no experiments were done on LLM fine-tuning cases.

There are some recent explorations of using LLM for product attribute extraction \citet{fang2024llmensembleoptimallargelanguage, sabeh2024empiricalcomparisongenerativeapproaches, brinkmann2024productattributevalueextraction}. Different from their focus, our work mainly explores effective fine-tuning strategies and product attributes extraction is the specific task to verify the effectiveness of our approaches. Moreover, in our experiments, we not only tested it in product attributes but also other generic information such as store contact, seller information, etc. 

\section{Fine-tuning Task Description and Dataset Construction}

\subsection{Fine-tuning Task Description}
The primary fine-tuning task in this paper is to steer the LLM to accurately extract and classify specific attribute values from a diverse array of e-commerce web pages. This task is critical in enhancing the model's applicability to real-world e-commerce scenarios, where accurate information extraction directly influences business outcomes and customer experiences.

Our training data is sourced from e-commerce website, as raw HTML and URLs. We use LLM queries to extract the following three types of information: First, the class of web page (contact page, product details, list of products, etc). Second, a set of attribute values. For example, for product details pages, we extract attributes such as the title, price and description. Third, forms URLs we are interested in from certain types of pages, such as keyword search JavaScript blocks.

\subsection{Dataset Construction}
We define a training sample of 5000 web pages, re-weighted to sample rarer classes of web pages such as contact pages. 
For this study, we focus on 8 attributes, 4 found in product details pages and 4 found in contact pages, listed in Table \ref{tab:attribute_table} in Appendix \ref{sec:data_information}. We extract the "ground truth" with OpenAI's GPT4 \cite{openai2024chatgpt}. As the set of attributes we query for a web-page depends on the classification, the accuracy of the extraction depends partially on the accuracy of the web-page classification. For each page, we use our "truth" LLM, GPT4, to do each of the above tasks, saving the prompt, input, and output. We then filter and correct a few common mistakes made by GPT4. From the 5000 pages and the three types of queries, we produced 12k sets of prompts, inputs, and outputs, which is roughly equivalent to $100M$ tokens.


In order to evaluate the accuracy of our models on our prioritized page types (product details and contact pages), we compile and label (with GPT4) two test sets. The first test set of 200 pages (100 product details and 100 contact pages) serves as a validation set, being used to test model accuracy in both the data efficiency and hyperparameter optimization studies. In both studies, choosing the sample sizes or hyperparameters with the best accuracy on the first test set allows the models to over-fit the testing set. For this reason, we define a second test test of 242 product details pages and 242 contact pages, to determine the final accuracy of our best models.

\section{Data Efficiency Analysis}
\label{sec:data_efficiency}

To investigate the relationship between fine-tuning data volume and model performance, we conduct a systematic study varying the number of training samples while maintaining a consistent set of hyperparameters given in Table \ref{tab:data_eff_hyperparameters}. Our experimental design encompasses a range of 0 to 10,000 samples, with the original Llama-3-8B-Instruct model serving as our baseline (N = 0).


\begin{table}[ht]
\centering
\begin{tabular}{ll}
\hline
\textbf{Training Hyperparameters} & \textbf{Values} \\ 
\hline
Epochs            & 12        \\ 
Learning Rate     & 0.0001    \\ 
Batch size        & 16        \\ 
LoRA targets      & $q_{proj}$, $v_{proj}$ \\ 
LoRA Rank         & 16        \\ 
LoRA Alpha        & 32        \\ 
LoRA Dropout      & 0.1       \\ 
LR Schedule       & cosine    \\ 
LR Warmup         & 0.1       \\ 
\hline
\end{tabular}
\caption{Training Hyperparameters}
\label{tab:data_eff_hyperparameters}
\end{table}

\subsection{Accuracy Metrics}

We define attribute-specific accuracy as the ratio of correctly extracted non-empty attribute values to the total number of non-empty values in the ground truth set. This metric accounts for the variable presence of attributes across web pages while not penalizing hallucinations in cases of empty truth values. Our overall accuracy metric is the mean of eight individual attribute accuracies, equally weighted between product details and contact page attributes.

\begin{figure}[ht]
    \centering
    \includegraphics[width=\linewidth]{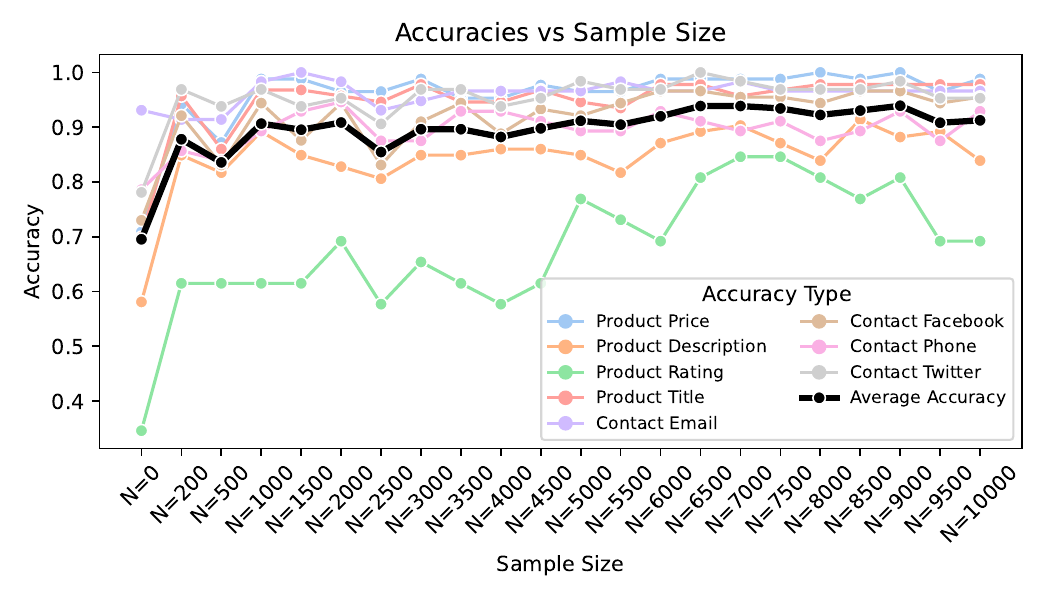}
    \caption{Accuracy for different attributes and average accuracy, for different amounts of training data.}
    \label{fig:accuracies_vs_data}
\end{figure}

We picked 4 attributes for product details pages and 4 attributes for contact pages to evaluate, based on our interest and their relative frequency. For each model, we compute an average of these 8 accuracies, and use this as our general metric for the models. The individual accuracies and the average accuracies for our models are shown in figure \ref{fig:accuracies_vs_data}.

\subsection{Discussion}

Our findings reveal several key insights:

\begin{itemize}
    \item Rapid Initial Improvement: Even with a modest 200 samples (approximately 100 web pages), we observe a substantial increase in model accuracy from 70\% to 88\%.
    \item Diminishing Returns: The majority of accuracy gains are achieved by 1,000 samples, with subsequent improvements becoming more gradual.
    \item Attribute-Specific Trends: Late-stage accuracy improvements are predominantly driven by a single attribute type - product rating - which is notably less frequent in our dataset, appearing in only about 25\% of product details pages.
    \item Performance Plateau: Maximum performance is generally attained at approximately 6,500 samples, suggesting a potential "sweet spot" for data efficiency in our specific task domain.
\end{itemize}

These results underscore the importance of strategic data sampling in fine-tuning large language models for specialized tasks. While even small datasets can yield significant improvements, careful consideration must be given to the distribution of attributes within the training data to ensure comprehensive model performance across all target variables.
Our findings have important implications for practitioners seeking to optimize the data collection and annotation process for fine-tuning LLMs in resource-constrained environments. Future work could explore the impact of data augmentation techniques or alternative sampling strategies to further enhance model performance, particularly for less frequent attributes.




\section{Hyperparameter Optimization}
\label{sec:hyper_parameter_optimization}

\subsection{Problem Formulation}

The task of efficient hyperparameter optimization can be formalized as a two-part optimization problem:

\begin{itemize}
    \item Finding the optimal set of hyperparameters $\theta^*$ that maximizes the performance metric $M$ of a model $f$ on a validation set $D_{val}$:
$$\theta^* = argmax_{\theta} M(f_{\theta}(D_{val}))$$
where $f_{\theta}$ represents the model trained with hyperparameters $\theta$. In our context, $M$ is the average accuracy across eight selected attributes.
    \item Maximize the correlation $\mu$ between the model's performance at early step $t_1$ and its performance at final step $t_2$:
    $$max \ \mu (M(f_{\theta}(D_{val}, t_1)), M(f_{\theta}(D_{val}, t_2)))$$, where $t_2$ represents the end of the training process (in our case, 80\% of total epochs).
\end{itemize}

This formulation encapsulates our goal of developing an efficient tuning method that uses less computation time by leveraging early-stage performance ($t_1$) to predict final performance ($t_2$), while still achieving high final accuracy.

\subsection{Methodology}

\setlength{\textfloatsep}{5pt} 
\setlength{\intextsep}{5pt}    
\setlength{\floatsep}{5pt}     

\begin{algorithm}
\setlength{\baselineskip}{8pt} 
\caption{Efficient Hyperparameter Optimization}
\begin{flushleft}
\textbf{Input:} Hyperparameter space $\Theta$, iterations $N$, early eval time $t_1$, final eval time $t_2$ \\
\textbf{Output:} Optimal hyperparameters $\theta^*$
\begin{enumerate}
    \item Initialize results pool $P \gets \emptyset$
    \item \textbf{for} $i = 1$ to $N$ \textbf{do}
    \item \quad $\kappa \gets \kappa_{\text{explore}}$ if $i \leq N/2$, else $\kappa_{\text{exploit}}$
    \item \quad $\theta_i \gets \text{BayesianOptimization}(P, \kappa)$
    \item \quad Train model $f_{\theta_i}$ up to time $t_1$
    \item \quad Evaluate $M_i = M(f_{\theta_i}(D_{\text{val}}, t_1))$
    \item \quad $P \gets P \cup \{(\theta_i, M_i)\}$
    \item Select top $k$ configurations from $P$
    \item \textbf{for} each $\theta_j$ in top $k$ configurations \textbf{do}
    \item \quad Continue training $f_{\theta_j}$ from $t_1$ to $t_2$
    \item \quad Evaluate $M_j = M(f_{\theta_j}(D_{\text{val}}, t_2))$
    \item $\theta^* \gets \text{argmax}_{\theta_j} M_j$
    \item \textbf{return} $\theta^*$
\end{enumerate}
\end{flushleft}
\end{algorithm}

The motivating idea behind this is to implement a version where we fine-tune models only until time $t_1$ and freeze the fine-tuning, running hyperparameter optimization with a Bayesian procedure until finding an ideal set of hyperparameters, and unfreezing the models with those hyperparameters. In order to test whether such a method would work, we fine-tune for the whole duration, but use the model generated early at time $t_1$ for evaluation for the hyperparameter optimization. We defined $t_1$ in terms of the fraction of the total number of epochs, so that the amount of data seen by the model by time $t_1$ does not depend on the batch size. We then define models at two later times, $t_2$ and the minima of the validation loss that is computed by LLaMA-Factory. For this study, we set $t_1=0.2$ and $t_2=0.8$. By checking if the best models at $t_2$ and the minimum loss time are also the best models at time $t_1$, we can determine if the proposed freezing and hyperparameter optimization procedure is viable. If not, that would indicate the hyperparameters optimizing the performance at $t_1$ are not necessarily the best hyperparameters for the whole training time.

We design the following steps for the implementation of the hyperparameter optimization:

\begin{enumerate}
    \item Run the LoRA fine-tuning a set of hyperparameters
    \item Evaluate the accuracy on a validation test set using the model at step $t_1$
    \item Save the hyperparameter configuration and accuracy to pool of all results
    \item Run a Bayesian optimization algorithm on the pool to produce the next set of hyperparameters
\end{enumerate}


At each step, optimizer updates a Gaussian process regression surrogate model of the black-box system, which is where the function mapping between input hyperparameter sets and output average model accuracies (technically the negative accuracies, as scikit-optimize only has a "minimize" function, and no "maximize" function). We used the lower confidence bound (LCB)~\citet{} acquisition function, which picks new configurations to check using the surrogate model and its mean $\mu(x)$ and variance $\sigma(x)$ models according to:

\begin{equation}
    f(x)=\mu(x) - \kappa \sigma(x)
\end{equation}

The recommended configurations are found from the minima of $x$ in $f(x)$. When the $\kappa$ value is high, this prioritizes the exploration of configuration space areas with low uncertainty. When it is low, this prioritizes the exploration of areas of low mean and low uncertainty. We incorporate this into our loop by having the instances use the high value of $\kappa$, $\kappa_{\text{explore}}$ during the first half of the experiment, and then a lower value, $\kappa_\text{exploit}$ in the second half. The specific values for these, as well as the other global parameters of our study, are given in Table \ref{tab:opt_study_global_parameters} in Appendix \ref{sec:hpo_tables}.

\subsection{Hyperparameter Space}

Our study focuses on optimizing 6 key hyperparameters of the LoRA fine-tuning process. These hyperparameters were selected based on their significant impact on model performance and their potential for interaction effects.
The first set was \emph{LoRA target layers}. Two standard approaches are to adapt the $q_{proj}$ and $v_{proj}$ matrices or to adapt $q_{proj}$, $k_{proj}$, $v_{proj}$, and $o_{proj}$ matrices \cite{hu2021lora}. To give a richer parameter space to explore the trade-off between adaptation flexibility and computational efficiency, we define 4 choices of target layers, indexed by an integer that adds a new layer with each increase. 

In addition to the LORA layers, we include the learning rate, the batch size, (To keep memory usage under control, we used gradient accumulation to simulate batch sizes.), and three more hyperparameters of LoRA: the rank $R$, LORA $\alpha$, and the dropout. The rank $R$ determines the dimensionality of the low-rank approximation in LoRA. Higher $R$ increase model capacity but also computational cost. The LoRA $\alpha$ parameter sets the strength of the LoRA adapter matrix relative to the pretrained projection matrices and is similar to the learning rate \cite{hu2021lora}. Finally, the LoRA dropout parameter controls the probability for the dropout mechanism in LoRA, described in \cite{lin2024loradropoutsparsityregularizer}, masks rows and vectors from the LoRA matrices. 

Two hyperparameters that we do not vary are the choice of LR scheduler and the LR warm-up period, for which we use the LLaMA-Factory defaults of \texttt{cosine} and 0.1, respectively.

The ranges we use for each of the hyperparameters are given in Table \ref{tab:hpo_parameters}.

\begin{table}[ht]
    \centering
    \begin{tabular}{c|c|c|c}
        \textbf{Hyperparameter} & \textbf{Type} &  \textbf{Min} & \textbf{Max} \\ \hline
        LoRA Target Index & Integer & 0 & 3 \\
        Learning Rate & Float & 0.00001 & 0.01 \\
        Batch Size & Integer & 1 & 32 \\
        LoRA Rank & Integer & 4 & 64 \\
        LoRA Alpha & Float & 0.1 & 128 \\
        LoRA Dropout & Float & 0.1 & 0.8 \\
    \end{tabular}
    \caption{Hyperparameters to optimize and the ranges used in our study.}
    \label{tab:hpo_parameters}
\end{table}

\subsection{Results and Analysis}

\subsubsection{Performance Correlation Across Training Stages}

Our study reveals a strong correlation between early-stage model performance and final model accuracy, validating our hypothesis that early evaluation can effectively predict overall model quality. This finding has significant implications for efficient hyperparameter optimization in large language model fine-tuning.

\begin{figure}
    \centering
    \includegraphics[width=1.0\linewidth]{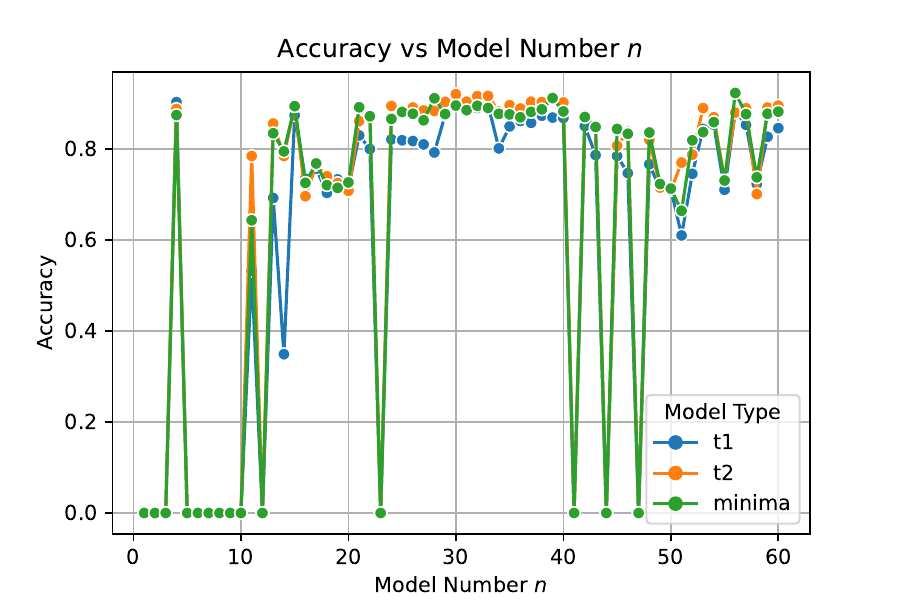}
    \caption{Accuracy score vs model number $n$. The accuracy of the models produced at $t_1$, $t_2$, and at the evaluation minima are shown together. In the early stages, where the study is in the exploration phase, more of the tested models have 0 accuracy. Only 3 out of 30 models failed in the exploitation phase.}
    \label{fig:accuracy_vs_n}
\end{figure}

To test if the models produced by tuning to the accuracy at time $t_1$ are also optimized models at $t_2$ and at the minimum loss time, we test the validation accuracy of each of the models with different hyperparameter settings derived from the Bayesian optimization phase. The total of 60 models' performance is shown in figure \ref{fig:accuracy_vs_n}. Models with 0 accuracy indicate that the fine-tuning produced non-functioning models, the details of this are investigated in Appendix \ref{sec:hpo_further_disc}. From there, we see early performance as a good indicator: Of the top 5 models at time $t_1$ (20\% of total training time), 4 remained in the top 5 at time $t_2$ (80\% of total training time). This remarkable consistency demonstrates the predictive power of early-stage evaluation~\footnote{The top 10 models at each time are given in table \ref{tab:best_hpo_models} in Appendix \ref{sec:hpo_tables}}.

\emph{Correlation with Loss Minima}. By chance, one of the randomly chosen initial models (model 4) performs the best at time $t_1$, but this model is not in the top 10 models at the minima or at $t_2$. The other 4 models in the top 5 $t_1$ models are all in the top 5 $t_2$ models, suggesting that the $t_1$ performance is a good predictor of the $t_2$ performance. However, only 2 of the top 5 $t_1$ models had corresponding minima models in the top 5. Of the top 5 minima models, 4 of the models had corresponding entries in the top 10 $t_1$ models. This suggests that while early performance is a strong indicator of final accuracy, the relationship with the loss minimum is more nuanced.

To mitigate potential overfitting concerns, we evaluate the top 5 models on a larger, independent test set, with the results in Table \ref{tab:hpo_models_best_extra_test} in Appendix \ref{sec:hpo_tables}. The results double confirm the strong correlation between early and final performance, with only minor shifts in relative rankings.

One final test is done to compare the best model found in the HPO procedure to the model created during the data efficiency study with the same sample size (although the data efficiency study used 12 epochs instead of 10 for the HPO study). This test is done with the independent test set of 242 product details and 242 contact pages, and the comparison is in Table \ref{tab:hpo_data_eff_comparison.}. The hyperparameter optimization improves the accuracy by about 2\% in comparison to the results found in the data efficiency study for the same number of samples.

\begin{table}[ht]
    \centering
    \begin{tabular}{l|c}
        Model ID & Accuracy \\\midrule
        Data Eff. (N=1000) &  0.8848 \\
        HPO (Model 56) & 0.9042 \\
    \end{tabular}
    \caption{Comparison of the best result from the HPO and the comparable model from the data efficiency study.}
    \label{tab:hpo_data_eff_comparison.}
\end{table}

\section{Conclusions}

Our study underscores the success of efficient fine-tuning strategies for Large Language Models, particularly using LoRA with only 200 samples to enhance product attribute value extraction significantly. Our hyperparameter optimization approach, which involves early model evaluation, effectively predicts final performance, confirming that early accuracy strongly correlates with later outcomes. This method offers a more resource-efficient way of tuning, which is beneficial for practitioners aiming to refine LLM fine-tuning while conserving resources and ensuring high performance.

\bibliography{custom}

\bibliographystyle{acl_natbib}

\appendix

\section{Attribute Information}
\label{sec:data_information}

The attributes that we query and evaluate for accuracy for product and contact pages are listed in Table \ref{tab:attribute_table}.

\begin{table}[ht]
    \centering
    \begin{tabular}{c|c}
        Page Class & Attribute \\\hline
         Product & Price, Description, Rating, Title \\\hline
         Contact & Email, Facebook, Twitter, Phone
    \end{tabular}
    \caption{Target attributes of web pages.}
    \label{tab:attribute_table}
\end{table}

\section{Data Efficiency Tables}
\label{sec:data_eff_tables}

The accuracies for each of the models for each type of data are given in Table \ref{tab:date_eff_all}. 

\begin{table*}[ht]
\centering
\small
\begin{tabular}{p{1.1cm}|p{1.3cm}|p{1.3cm}|p{1.3cm}|p{1.3cm}|p{1.3cm}|p{1.3cm}|p{1.3cm}|p{1.3cm}|p{1.3cm}}
\toprule
 & \multicolumn{9}{c}{Accuracy} \\
\cmidrule{2-10}
Number of Samples & Price & Product Description & Product Rating & Product Title & Contact Email & Contact Facebook & Contact Phone & Contact Twitter & Average \\
\midrule
0 & 0.709 & 0.581 & 0.346 & 0.699 & 0.931 & 0.730 & 0.786 & 0.781 & 0.695 \\
200 & 0.942 & 0.849 & 0.615 & 0.957 & 0.914 & 0.921 & 0.857 & 0.969 & 0.878 \\
500 & 0.872 & 0.817 & 0.615 & 0.860 & 0.914 & 0.831 & 0.839 & 0.938 & 0.836 \\
1000 & 0.988 & 0.892 & 0.615 & 0.968 & 0.983 & 0.944 & 0.893 & 0.969 & 0.906 \\
1500 & 0.988 & 0.849 & 0.615 & 0.968 & 1.000 & 0.876 & 0.929 & 0.938 & 0.895 \\
2000 & 0.965 & 0.828 & 0.692 & 0.957 & 0.983 & 0.944 & 0.946 & 0.953 & 0.908 \\
2500 & 0.965 & 0.806 & 0.577 & 0.946 & 0.931 & 0.831 & 0.875 & 0.906 & 0.855 \\
3000 & 0.988 & 0.849 & 0.654 & 0.978 & 0.948 & 0.910 & 0.875 & 0.969 & 0.896 \\
3500 & 0.953 & 0.849 & 0.615 & 0.946 & 0.966 & 0.944 & 0.929 & 0.969 & 0.896 \\
4000 & 0.953 & 0.860 & 0.577 & 0.946 & 0.966 & 0.888 & 0.929 & 0.938 & 0.882 \\
4500 & 0.977 & 0.860 & 0.615 & 0.968 & 0.966 & 0.933 & 0.911 & 0.953 & 0.898 \\
5000 & 0.965 & 0.849 & 0.769 & 0.946 & 0.966 & 0.921 & 0.893 & 0.984 & 0.912 \\
5500 & 0.965 & 0.817 & 0.731 & 0.935 & 0.983 & 0.944 & 0.893 & 0.969 & 0.905 \\
6000 & 0.988 & 0.871 & 0.692 & 0.978 & 0.966 & 0.966 & 0.929 & 0.969 & 0.920 \\
6500 & 0.988 & 0.892 & 0.808 & 0.978 & 0.966 & 0.966 & 0.911 & 1.000 & 0.939 \\
7000 & 0.988 & 0.903 & 0.846 & 0.957 & 0.983 & 0.955 & 0.893 & 0.984 & 0.939 \\
7500 & 0.988 & 0.871 & 0.846 & 0.968 & 0.966 & 0.955 & 0.911 & 0.969 & 0.934 \\
8000 & 1.000 & 0.839 & 0.808 & 0.978 & 0.966 & 0.944 & 0.875 & 0.969 & 0.922 \\
8500 & 0.988 & 0.914 & 0.769 & 0.978 & 0.966 & 0.966 & 0.893 & 0.969 & 0.930 \\
9000 & 1.000 & 0.882 & 0.808 & 0.978 & 0.966 & 0.966 & 0.929 & 0.984 & 0.939 \\
9500 & 0.965 & 0.892 & 0.692 & 0.978 & 0.966 & 0.944 & 0.875 & 0.953 & 0.908 \\
10000 & 0.988 & 0.839 & 0.692 & 0.978 & 0.966 & 0.955 & 0.929 & 0.953 & 0.912 \\
\bottomrule
\end{tabular}
\caption{The accuracy for models trained on each data size and attribute. The average of the accuracies for all attributes is included.}
\label{tab:date_eff_all}
\end{table*}

\clearpage

\section{Hyperparameter Optimization Notes}
\label{sec:hpo_notes}

\subsection{Zero Accuracy Models}
\label{sec:hpo_further_disc}
\begin{figure}
    \centering
    \includegraphics[width=1.0\linewidth]{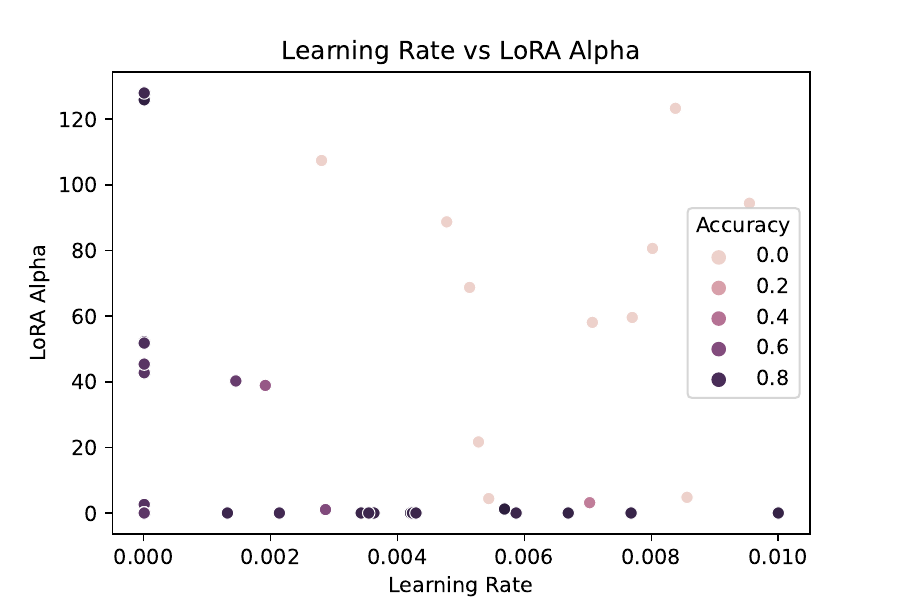}
    \caption{Learning Rate vs LoRA $\alpha$ with resulting model accuracies. Models with high values for both parameters had significant fitting issue, producing useless models with 0 accuracy.}
    \label{fig:LR_vs_alpha_accuracy}
\end{figure}

One notable finding during the experiment is that some combinations of hyperparameters produced gibberish without end tokens, which added to the testing time. For these models, we assign an accuracy of 0. It was found that these issues occurred when both the LoRA $\alpha$ and learning rate were high, as visible in Figure \ref{fig:LR_vs_alpha_accuracy}. This is understandable in light of the close relationship between learning rate and alpha described by \cite{hu2021lora}. Our optimization procedure automatically detects this relationship, and does not use a high $\alpha$ parameter and learning rate together during the exploitation phases.


\section{Hyperparameter Optimization Tables and Additional Plots}
\label{sec:hpo_tables}

\begin{figure}[ht]
    \centering
    \includegraphics[width=1.0\linewidth]{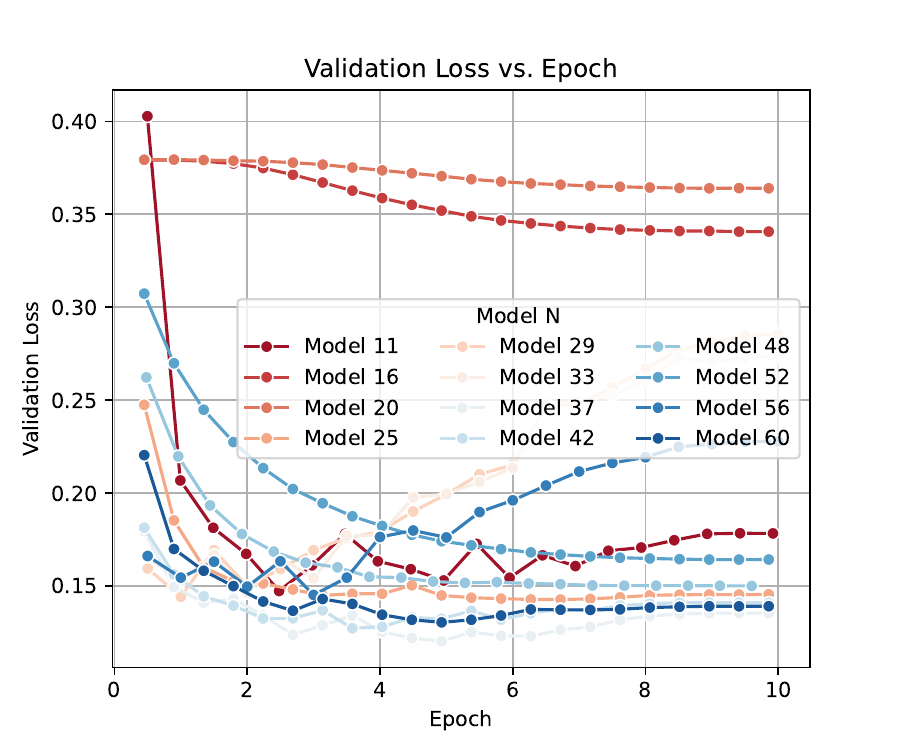}
    \caption{Validation Loss vs Epoch for a sample of models. Models for which the training failed significantly are excluded.}
    \label{fig:eval_loss_plots_1}
\end{figure}

The validation loss curves for a sample of the models, excluding those that failed significantly in training, is shown in Figure \ref{fig:eval_loss_plots_1}.

Several parameters, including the data size, number of epochs, and Bayesian optimization parameters, in our hyperparameter optimization study are given in Table \ref{tab:opt_study_global_parameters}.

\begin{table}[ht]
    \centering
    \small
    \begin{tabular}{c|p{0.5\linewidth}|c}
        Parameter & Description & Value \\\midrule
        $t_1$ & Early model evaluation time & 0.2 \\
        $t_2$ & Late model evaluation time & 0.8 \\
        $N_{samples}$ & Number of samples & 1000 \\
        $N_{epochs}$ & Number of epochs & 10 \\
        \texttt{acq\_func} & Acquisition function for Bayesian optimization & \texttt{LCB} \\
        $\kappa_{\text{explore}}$ & Explore-exploit parameter for exploration phase & 5.0 \\
        $\kappa_{\text{exploit}}$ & Explore-exploit parameter for exploitation phase & 1.0 \\
    \end{tabular}
    \caption{Optimization Study Global Parameters}
    \label{tab:opt_study_global_parameters}
\end{table}

The top 10 model IDs and validation accuracies from each time ($t_1$, $t_2$, and the minima of the evaluation loss) are shown in Table \ref{tab:best_hpo_models}. These accuracies, especially the $t_1$ accuracies, are potentially affected by overfitting.

\begin{table*}[t]
    \centering
    \small
    \begin{tabular}{c|c||c|c||c|c}
        $t_1$ Model ID & $t_1$ Accuracy & $t_2$ Model ID & $t_2$ Accuracy & Minima Model ID & Minima Accuracy \\\midrule
        4 &             0.9022 &         30 &             0.9196 &         56 &             0.9222 \\
        30 &             0.8962 &         33 &             0.9159 &         39 &             0.9111 \\
        33 &             0.8961 &         32 &             0.9154 &         28 &             0.9106 \\
        31 &             0.8949 &         39 &             0.9138 &         30 &             0.8948 \\
        32 &             0.8894 &         31 &             0.9031 &         32 &             0.8941 \\ \midrule
        56 &             0.8829 &         37 &             0.903  &         15 &             0.8934 \\
        29 &             0.8741 &         29 &             0.9028 &         21 &             0.891  \\
        38 &             0.8724 &         38 &             0.9024 &         33 &             0.8894 \\
        39 &             0.8682 &         40 &             0.9015 &         38 &             0.8864 \\
        40 &             0.8674 &         35 &             0.8954 &         31 &             0.8846 \\

    \end{tabular}
    \caption{Best models from the HPO experiment with their accuracy on the validation data that was used for the optimization. The model ID corresponds to a unique training process, while the $t_1$, $t_2$, and minima models come from different points in each process. }
    \label{tab:best_hpo_models}
\end{table*}

The results from testing the accuracy of our best HPO models on an independent test set of data not used for the Bayesian optimization are given in Table \ref{tab:hpo_models_best_extra_test}. The Model IDs correspond roughly to the order in which they started fine-tuning.

\begin{table}[ht]
    \centering
    \small
    \begin{tabular}{c|c|c}
    Model Time & Model ID & Accuracy \\\midrule
    $t_1$ & 4 & 0.8791 \\
    $t_1$ & 30 & 0.8724 \\
    $t_1$ & 33 & 0.8632 \\
    $t_1$ & 31 & 0.9014 \\
    $t_1$ & 32 & 0.8742 \\\midrule
    $t_2$ & 30 & 0.894 \\
    $t_2$ & 33 & 0.8772 \\
    $t_2$ & 32 & 0.8828 \\
    $t_2$ & 39 & 0.8804 \\
    $t_2$ & 31 & 0.8921 \\\midrule
    Minima & 56 & 0.9042 \\
    Minima & 39 & 0.894 \\
    Minima & 28 & 0.8899 \\
    Minima & 30 & 0.8638 \\
    Minima & 32 & 0.8735 \\

\end{tabular}

    \caption{Best models from the HPO experiment, re-evaluated on an test set that is independent from the optimization. The models are listed in order of their scores on the validation test set used during the HPO.}
    \label{tab:hpo_models_best_extra_test}
\end{table}

The hyperparameter set for the top 10 models (evaluated at time $t_1$ are included in Table \ref{tab:best_hyperparameters}. Many sets of hyperparameters are repeated, this occurs when a parallel instance of the optimizer, after finishing a test and adding data new point, still picks the same set of hyperparameters to try next. This could be at least partially avoided by generating multiple sets of hyperparameters at a time. While this repetition is not our original intention, it does have the benefit of adding more precision to the testing of a hyperparameter set that is consistently viewed by the model as of interest. 

\begin{table}[ht]
    \tiny
    \centering
    \begin{tabular}{cccccccc}
        \toprule
        \thead{Model \\ N} & \thead{Accuracy} & \thead{LoRA \\ Target \\ Index} & \thead{Batch \\ Size} & \thead{Learning \\ Rate} & \thead{LoRA \\ Alpha} & \thead{LoRA \\ Dropout} & \thead{LoRA \\ Rank} \\
        \midrule
        4  & 0.9023  & 2 & 21 & 0.005687 & 1.304 & 0.7964 & 9 \\
        30 & 0.8963  & 3 & 1  & 1E-05                & 128.0              & 0.1                 & 64 \\
        33 & 0.8961 & 3 & 1  & 1E-05                & 128.0              & 0.1                 & 64 \\
        31 & 0.8949 & 3 & 1  & 1E-05                & 128.0              & 0.1                 & 64 \\
        32 & 0.8894 & 3 & 1  & 1E-05                & 128.0              & 0.1                 & 64 \\
        56 & 0.8829 & 1 & 1  & 1E-05                & 125.9 & 0.1811 & 59 \\
        29 & 0.8741 & 3 & 1  & 1E-05                & 128.0              & 0.1                 & 4  \\
        38 & 0.8734 & 0 & 32 & 0.01                 & 0.1                & 0.6618  & 64 \\
        39 & 0.8724 & 3 & 32 & 0.004205 & 0.1                & 0.1840 & 4  \\
        40 & 0.8683  & 3 & 32 & 0.01                 & 0.1                & 0.8                 & 4  \\
        \bottomrule
    \end{tabular}
    \caption{The top 10 hyperparameter sets identified at $t_1$.}
    \label{tab:best_hyperparameters}
\end{table}

\end{document}